\def\year{2018}\relax
\documentclass[letterpaper]{article} %DO NOT CHANGE THIS
\usepackage{amsmath}

\usepackage{aaai18}  %Required
\usepackage{times}  %Required
\usepackage{helvet}  %Required
\usepackage{courier}  %Required
\usepackage{url}  %Required
\usepackage{graphicx}  %Required

\usepackage{multirow}
\usepackage{tabularx}
\usepackage{graphicx}

\usepackage{amsfonts}       % blackboard math symbols
\usepackage{nicefrac}       % compact symbols for 1/2, etc.
\usepackage{microtype}      % microtypography
\usepackage{amsmath}
\usepackage{amsthm}
\usepackage{amssymb}
\usepackage{subfigure}

\begin{document}
% The file aaai.sty is the style file for AAAI Press 
% proceedings, working notes, and technical reports.
%
\title{Improving Variational Encoder-Decoders in Dialogue Generation}
\author{Xiaoyu Shen$^{1, 3}$\thanks{Authors contributed equally. Correspondence to  X. Shen (xshen@lsv.uni-saarland.de).}, Hui Su$^4\footnotemark[1]$, Shuzi Niu$^4$ and Vera Demberg$^{2,3}$\\
$^1$Max Planck Institute Informatics, Germany\\
$^2$Dept of Math \& CS and LangSci \& Tech, Saarland University, Germany\\
$^3$Saarland Informatics Campus, Germany\\
$^4$Institute of Software, University of Chinese Academy of Science, China\\
}
\maketitle
\begin{abstract}
Variational encoder-decoders (VEDs) have shown promising results in dialogue generation. However, the latent variable distributions are usually approximated by a much simpler model than the powerful RNN structure used for encoding and decoding, yielding the KL-vanishing problem and inconsistent training objective. In this paper, we separate the training step into two phases: The first phase learns to autoencode discrete texts into continuous embeddings, from which the second phase learns to generalize latent representations by reconstructing the encoded embedding.  In this case, latent variables are sampled by transforming Gaussian noise through multi-layer perceptrons and are trained with a separate VED model, which has the potential of realizing a much more flexible distribution. We compare our model with current popular models and the experiment demonstrates substantial improvement in both metric-based and human evaluations.
\end{abstract}

\section{Introduction}
Recurrent neural networks (RNNs)~\cite{bengio2003neural} are widely used in natural language processing tasks. However, given the history context, RNNs estimate the probability of one word at a time and does not work from a holistic sentence representation~\cite{bowman2016generating}. When applied to dialogue generation, the corresponding result is that it would generate either short, boring responses or long, inconsistent sentences. As the length of generated sentences grows, it would easily deviate from the original intention as such token-level estimation only considers immediate short rewards and neglects global structure consistency. In hence, vanilla RNNs prefer generating generic and safe short responses to avoid the risk of making errors~\cite{vinyals2015neural,serban2015building,shen2017conditional}. One way of improving this deficient generating process is to introduce a sentence-level representation, which can be further conditioned on to ensure the sentence-level consistency.

Deep latent variable models are a popular way to learn such representations in a generative setting.  Latent representations and generators can be jointly trained in an unsupervised way. By learning the probability of synthesizing real data from intermediate latent variables, they are expected to uncover and disentangle causal factors that are most important to explain the data. The exact log-likelihood normally requires integral in high-dimensional space and cannot be analytically expressed. Current approaches solve this intractability problem by imposing a recognition network to approximate the real posterior probability. Variational autoencoders (VAEs)~\cite{kingma2013auto,rezende2014stochastic} bring scalability and stability to the training procedure, which introduces a reparameterization trick to reduce the variance when estimating the backpropagated gradients.

\cite{serban2016hierarchical} proposed the VHRED structure which applied the conditional VAE (CVAE)~\cite{sohn2015learning} with RNN encoder-decoders in dialogue generation, in hope of CVAE's advantage of learning global representations being a good complement of RNN's power at modeling local dependencies. However, this simple combination runs into the KL-vanishing problem that the RNN part ends up explaining all the structures without making use of the latent representation. The reason is that RNN is a universal approximator with much more flexibility than the simple gaussian distributed latent variables so that the model lacks enough motivation to utilize them. 

Current approaches normally address this problem by weakening the RNN decoder to match the simpler latent variable distribution, which essentially sacrifices the generating capacity for better representation learning and is inappropriate when our main goal is to learn a generative model. In this paper, on the contrary, we take advantage of the universality of RNNs to help realize a more flexible latent variable distribution. By this means, we can not only add motivation for utilizing latent variables, but also strengthen the expressiveness of the generating model. Specifically, we split the whole structure into a CVAE module and an autoencoder (AE) module. The CVAE module learns to generate latent variables while the AE module builds the connection between them and real dialogue utterances. The outputs of the CVAE serve as input latent variables for the AE module, which is potentially much more flexible than restricting the latent variables to follow a fixed distribution. As the RNN encoder-decoders in the AE module are universal approximators, they are adjusted to extract continuous vectors from the dialogue data that can be more easily modelled by the CVAE module. Combined with a scheduled sampling trick, this structure can significantly improve the generating performance. We show this structure can be compared to an adversarial encoder-decoder which substitutes the GAN step with a VAE alternative. Though theoretically less accurate, our framework is preferred to AED as the training process of VAE is much more reliable than GAN in seq2seq tasks and the universality of RNN ensures this inaccuracy can be controlled within an acceptable range.

\section{VED in Dialogue Generation}
In this section, we review the VAE and VHRED structure, then analyze where the training difficulty comes from when applied in dialogue generation and how current approaches try to solve this problem.
\subsection{VAE and VHRED}
The variational autoencoder (VAE)~\cite{kingma2013auto,rezende2014stochastic} is a popular generative model. Its generating process is as follows: data $x$ is generated by the generative distribution $p_\theta(x|z)$ and $z$ is sampled from the prior distribution $p(z)$. In contrast to calculating the exact log-likelihood, it can be efficiently trained by optimizing a valid lower bound~\cite{jordan1999introduction}. The objective takes the following form:
\begin{align}
\label{eq: vae}
\begin{split}
	&-\log  p_{\theta}(x) \leq -\log p_\theta(x)+\text{KL}(q_\phi(z|x)||p_\theta(z|x)) \\
  &= -\mathbb{E}_{q_{\phi}(z|x)} [\log p_\theta(x|z)]
  	+ \text{KL}(q_{\phi}(z|x || p(z)) 
\end{split}
\end{align}
$p_\theta(z|x)$ is the real posterior distribution of $z$ given the prior distribution $p_\theta(z)$ and the likelihood $p_\theta(x|z)$. The optimizing objective is namely maximizing the likelihood $\log p_\theta(x)$ and at the same time minimizing the mismatch between the approximated posterior $q_\phi(z|x)$, which is parametrized by neural networks, and the real posterior $p_\theta(z|x)$. When the gap $\text{KL}(q_\phi(z|x)||p_\theta(z|x))$ is large, the objective becomes inconsistent and the generating process cannot recover the real data distribution even in the global optimum.

The whole process can be conditioned on an additional context $c$, which leads to the conditional VAE~\cite{sohn2015learning} (CVAE): the output $x$ is generated from the distribution $p_\theta(x|c,z)$, latent variable $z$ is drawn from the prior distribution $p_\theta(z|c)$. The variational lower bound of CVAE is written as follows:
\begin{align}
\label{eq: cvae}
\begin{split}
	-\mathbb{E}_{q_\phi (z|x,c)} [\log p_\theta(x|c,z)]
                       +  \text{KL}(q_\phi (z|x,c) \| p_\theta (z|c))
\end{split}
\end{align}                     
Specially, to some extent, when both the context $c$ and output $x$ are sequential data, CVAE can also be treated as a seq2seq model~\cite{sutskever2014sequence}.

The variational hierarchical recurrent encoder-decoder (VHRED)~\cite{serban2016hierarchical} is a CVAE with hierarchical RNN encoders, where the first-layer RNN encodes token-level variations and the second-layer RNN captures sentence-level topic shifts. In this case, $c$ in Equation. \ref{eq: cvae} stands for dialogue history, $x$ is the response to be decoded and $z$ is the latent variable reflecting the high-level representation of $x$. The distribution $q_\phi(z|x,c)$ and $p_\theta(z|c)$ are usually set as simple Gaussian distributions with diagonal covariance matrix.

\subsection{Optimization Challenges}
\label{sec: why}
In VHRED, straightforwardly optimizing with Equation. \ref{eq: cvae} suffers from the KL-vanishing problem because the RNN decoder $p_\theta(x|c,z)$ is a universal function approximator and tends to represent the distribution without referring to the latent variable. At the beginning of the training process, when the approximate posterior $q_\phi(z|x,c)$ carries little useful information, it is natural for the model to blindly set $q_\phi(z|x,c)$ closer to the Gaussian prior $p_\theta(z|c)$ so that the extra cost from the KL divergence can be avoided~\cite{chen2016variational}.

To better analyze where the optimizing comes from, we can rewrite Equation. 2 as the following:
\begin{align}
\label{eq: why-cvae}
\begin{split}
-\log \int_{z} p_\theta(z|c) p_{\theta}(x|z,c) dz+\text{KL}(q_\phi(z|x,c)||p_\theta(z|x,c))
\end{split}
\end{align}

Let's first take a look at the first item, $\log \int_{z} p_\theta(z|c) p_{\theta}(x|z,c) dz = \log p_\theta(x|c)$. When the family of $p_\theta(x|z,c)$ is complex enough and includes the real distribution of $x$, the optimal value of this item is $p(x|c)$ and the reliance on $z$ is not necessary. However, reliance on $z$ provides the model with a chance of taking advantage of $z's$ distribution and reduces the complexity requirement for the distribution family $p_\theta(x|z,c)$. For example, suppose $p(x|c)=\mathcal{N}(0,1)$ and $p_\theta(z|c)=\mathcal{N}(3,1)$, modeling $p(x|c)$ accurately without reliance on $z$ requires $p_\theta(x|z,c)$ to include the Gaussian distribution, while by means of the linear mapping between $x$ and $z$, $p_\theta(x|z,c)$ can describe the real distribution with only linear complexity. When Gaussian distribution is not covered in the family $p_\theta(x|z,c)$, this model has to exploit the relation between $x$ and $z$ to model the real distribution. Likewise, in dialogue generation, although the RNN decoder $p_\theta(x|c)$ can in theory approximate arbitrary function, perfectly fitting the real dialogue distribution is still difficult due to the optimizing challenge, training corpus size and approximating errors. Therefore, to achieve the global optimum, we believe this first item will always prefer utilizing the latent variables, so long as the decoder $p_\theta(x|z,c)$ is not perfect. The weaker the decoder family is, the more it will be biased to utilizing latent variables. A more flexible prior distribution $p_\theta(z)$ will also increase the chance as it provides more possibilities for the utilisation.

The second item is the KL divergence, whose minimum value is 0 if and only if $q_\phi(z|x,c)=p_\theta(z|x,c)$. According to the Bayes theorem, we can express $p_\theta(z|x,c)$ as:
\begin{equation}
p_\theta(z|x,c)=\frac{p_\theta(x|z,c)p_\theta(z|c)}{p_\theta(x|c)}
\end{equation}
By ignoring the latent variable $z$, $p_\theta(x|z,c)$ and $p_\theta(x|c)$ cancels out, setting $q_\phi(z|x,c)=p_\theta(z|c)$ can easily arrive at the global optimum 0. Otherwise, when $p_\theta(z|c)$ is parametrised as a mean-field Gaussian distribution as in VHRED, the real posterior is impossible to fall into the same distribution family. Firstly, the independence relation cannot be satisfied. To make dimensions of $p_\theta(z|x)$ independent with each other, the likelihood $p_\theta(x|z)$ must exactly disentangle the effect of every dimension, which is unrealistic when $p_\theta(x|z)$ is a categorical distribution modelled by the RNN softmax. Secondly, the real posterior distribution can hardly still follow a Gaussian distribution when the likelihood $p_\theta(x|z)$ is based on discrete sequential data. Normally the training process will adjust $p_\theta(x|z)$ to make the real posterior easier to be modelled by $q_\phi(z|x)$~\cite{hinton1995wake}. However, when $x$ represents sentences with variable length, the value of $p_\theta(x|z)$ vanishes greatly when the length grows, which makes the adjusting task much more difficult. This implies the second item will always prefer ignoring the latent variables, so long as the approximated posterior is not powerful enough to perfectly match the real posterior. The weaker the approximating posterior distribution family is, the more it will be biased to ignoring latent variables.

Above all, the objective function of variational encoder-decoders in dialogue generation is essentially the competition of these two items, who is biased to utilizing or ignoring latent variables respectively. The reason of KL divergence vanishing in the global optimum is that the second term can gain more from ignoring the latent variables than the first term from utilizing them.

\subsection{Current Approaches}
If we use the ELBO objective, as explained, there are two directions to prevent the KL-vanishing problem: improving the advantage of utilising latent variables in $\log \int_{z} p_\theta(z|c) p_{\theta}(x|z,c) dz$ or weakening the advantage of abandoning latent variables in $\text{KL}(q_\phi(z|x,c)||p_\theta(z|x,c))$.

For the former direction, we need to use a smaller distribution family to model the decoder $p_\theta(x|z,c)$. When the decoder is weaker, if ignoring latent variables, it becomes farther from the real distribution at the global optimum thus encouraging latent variables to be exploited. Word drop-out~\cite{bowman2016generating} is a common method to weaken the RNN decoder. At each time step, the input word has a certain chance (drop-out rate) of becoming another word, the RNN decoder therefore cannot store a continuous history context. In \cite{xie2017data}, word drop-out is also explained as a special kind of smoothing. Similarly, for CNN decoders, limiting their power can also encode more information to latent variables~\cite{yang2017improved,chen2016variational}. Bag-of-word loss proposed by \cite{zhao2017learning} can also fall into this category. It imposes an extra loss which forces the latent variable to predict the whole sentence without word inputs, which is essentially increasing the weight of the reconstruction loss with the drop-out rate set to 1.

For the latter direction, we need to use a more flexible prior or posterior distribution for latent variables. Once the approximated posterior distribution is powerful enough, the KL divergence can be close to zero without losing the dependence on latent variables. \cite{serbanpiecewise} applies a piecewise distribution to replace the Gaussian prior distribution. Though can represent multi-modal conditions, it is still limited as a fixed distribution with pre-defined number of modes.  \cite{salimans2015markov} samples latent variables through Markov chains, but it imposed an extra approximation and the objective becomes less accurate. \cite{rezende2015variational,kingma2016improved,chen2016variational} use a normalizing flow. Latent variables are first sampled from a simple distribution then passed through several invertible transformations to get better flexibility. Normalizing flow is computationally more costly and has not been applied in text generation yet.

We can also change the original ELBO objective for easier optimization. KL-annealing~\cite{bowman2016generating} and free bits~\cite{kingma2016improved} are two popular strategies. In KL-annealing, a small weight is added to the KL divergence term in Equation. \ref{eq: cvae}, which starts from zero and gradually increases to 1. This prevents the model from zeroing out the KL divergence at the earlier training stage. Once the KL divergence vanishes, it is difficult to be recovered for the short sight nature of gradient descent. Free bits reserve some space of KL divergence for every dimension of latent variables. KL divergence is only optimized when exceeding the predefined quota. Similar ideas can be found in \cite{yang2017improved}, which reserved space for the total KL divergence instead of for every dimension.

\section{Improving Variational Encoder-Decoders}
\begin{figure*}[!ht]
\centering
\centerline{\includegraphics[height=9cm]{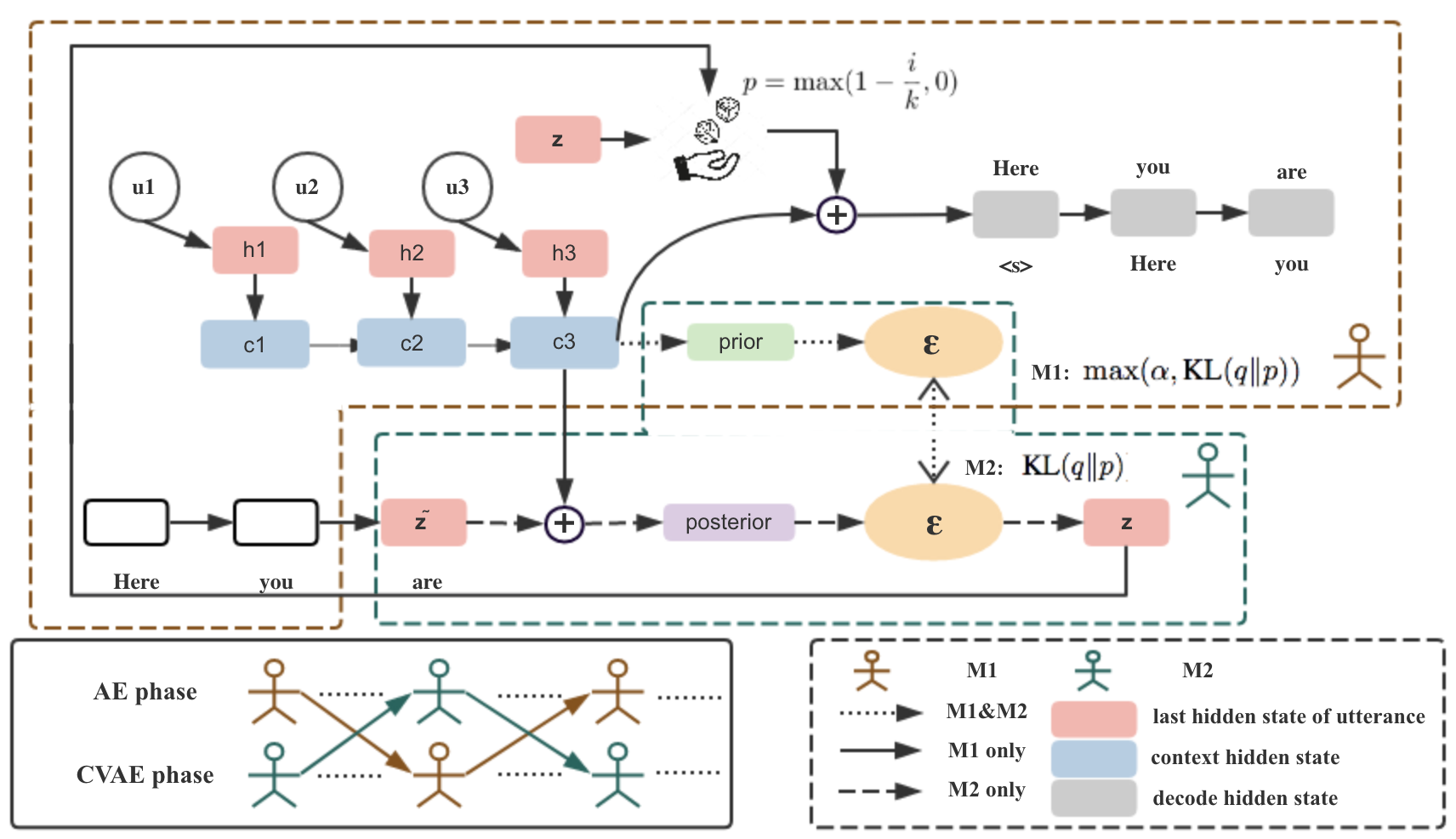}}
\caption{Architecture for collaborative variational encoder-decoder. $\bigoplus$ denotes concatenation of information. $M_1 (AE)$ and $M_2 (CVAE)$ are represented in brown and green respectively.}
\label{fig:model}
\end{figure*}
As discussed above, two ways for alleviating the optimizing challenge includes weakening the RNN decoders and improving the flexibility of latent variable distributions. The latter class is more fundamental since it also brings more expressiveness to the generating model. Weakening the decoders, though attenuating the KL-vanishing problem, will inevitably hurt the overall performance.

\subsubsection{Adversarial Encoder-Decoder}
An ideal way of representing the latent variable distribution is to use a universal approximator like neural networks. \cite{makhzani2015adversarial} proposed adversarial autoencoder (AAE) which samples posterior latent variables by transforming Gaussian noise through multi-layer-perceptrons. The flexibility of neural networks ensures it can fit arbitrary distribution. However, the probability density is intractable, so adversarial learning~\cite{goodfellow2014generative} must be implemented to replace the original KL divergence term.

\begin{figure}[!ht]
\centering
\centerline{\includegraphics[height=4.5cm]{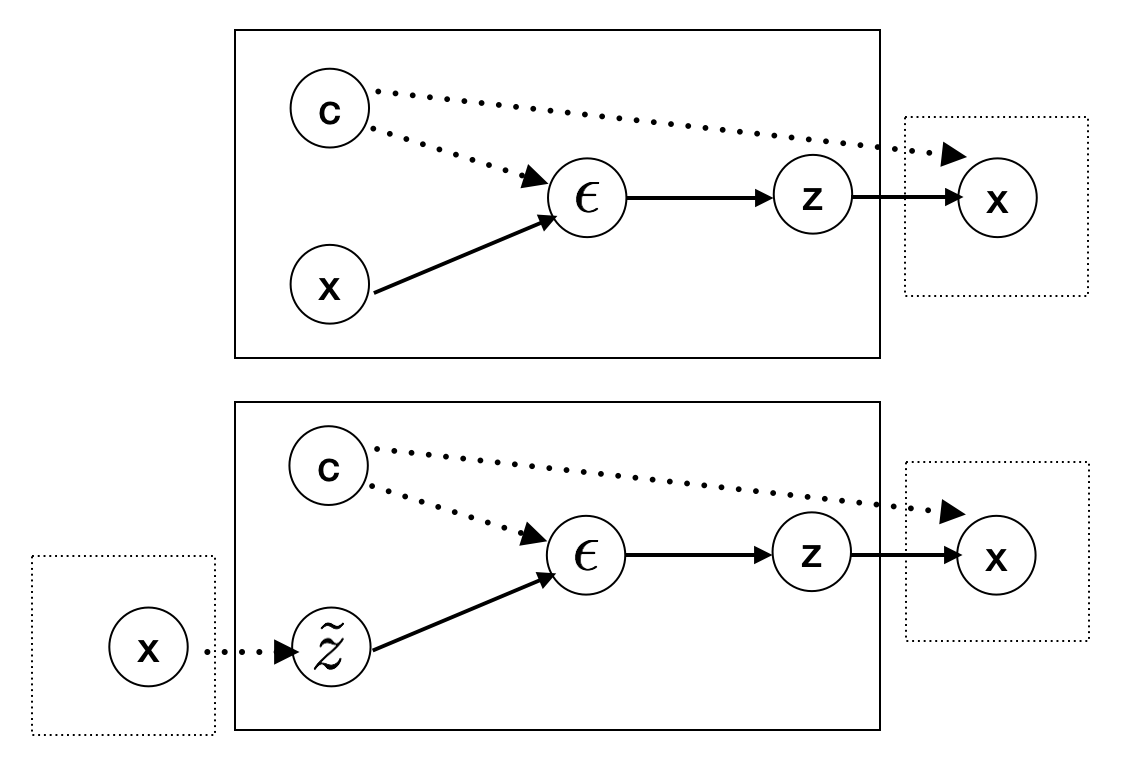}}
\caption{Up: adversarial encoder-decoder, down: adversarial encoder-decoder after replacing GAN with VAE, Full line rectangle: GAN and CVAE phase, Dotted line rectangle: AE phase}
\label{fig:connection}
\end{figure}

We can apply this idea to dialogue generation, where AAE is changed to context-dependent adversarial encoder-decoder (AED). The training objective can be represented as:
\begin{align}
\label{eq: aed}
\begin{split}
-\mathbb{E}_{q_\phi(z|c,x)}p_\theta(x|c,z)+JS(q_\phi(z|c)||p_\theta(z|c))
\end{split}
\end{align}
The training alternates between the autoencoder (AE) phase to optimize $-\mathbb{E}_{q_\phi(z|c,x)}p_\theta(x|c,z)$ and the GAN phase to match the aggregated posterior $q_\phi(z|c)$ and the prior $p_\theta(z|c)$. $q_\phi(z|c,x)$ and $p_\theta(z|c)$ are implicitly defined by passing context-dependent Gaussian random variables $\epsilon$ through multi-layer perceptrons. The graphical model is depicted in Figure. \ref{fig:connection}. It can be shown that this objective differs from the original ELBO by adding an extra punishment to the entropy of $q_\phi(x|z,c)$ and using Jensen-Shannon divergence in lieu of KL divergence. In the non-parametric limit, its generating model can recover the exact data distribution.
\subsubsection{Replacing GAN with VAE}
The idea of AED sounds appealing, but GAN is notoriously difficult to train, especially when both the prior and posterior need to be updated towards each other, the model becomes extremely sensitive to hyper-parameters and the training is very unstable. In consequence, we try replacing the GAN phase with a CVAE alternative. An RNN encoder is first applied to extract the corresponding latent variable target $\tilde{z}$ for each dialogue turn $x$, based on which a CVAE is trained to reconstruct it through context-dependent Gaussian noise. The connection to AED can be seen in Figure \ref{fig:connection}. Specifically, we just replace the $JS(q_\phi(z|c)||p_\theta(z|c))$ in Equation \ref{eq: aed} with the following CVAE objective:
\begin{align}
\label{eq: aed-cvae}
\begin{split}
-\mathbb{E}_{q_\phi(\epsilon|c,\tilde{z})}p_\theta(z|c,\epsilon)+KL(q_\phi(\epsilon|c,\tilde{z})||p_\theta(\epsilon|c))
\end{split}
\end{align}
$q_\phi(\epsilon|c,\tilde{z})$ is an approximated posterior. It can be easily proved when $q_\phi(\epsilon|c,\tilde{z})$ is powerful enough to cover the real posterior $p_\theta(\epsilon|c,\tilde{z})$, objective \ref{eq: aed-cvae} has the same global optimum as in $JS(q_\phi(z|c)||p_\theta(z|c))$. We can therefore instead alternate between the AE phase and the CVAE phase to achieve the same effect as in AED.
\subsubsection{Constraining RNN Encoder}
The accuracy of the CVAE objective relies on the matching degree of $q_\phi(\epsilon|c,\tilde{z})$ and $p_\theta(\epsilon|c,\tilde{z})$. Therefore, in the AE phase, apart from encoding representative information to reduce the normal AE reconstruction loss, the RNN encoder should also encode utterances in a manner where the real posterior $p_\theta(\epsilon|c,\tilde{z})$ can be more easily modelled by the distribution defined by $q_\phi(\epsilon|c,\tilde{z})$ in the CVAE phase. To do this, we add a KL divergence constraint to the RNN encoder in the AE phase. The RNN encoder has to keep $KL(q_\phi(\epsilon|c,\tilde{z})||p_\theta(\epsilon|c))$ within a specific range. It is also possible to constrain the value of the whole CVAE objective of Equation. \ref{eq: aed-cvae}, but we find constraining only the KL divergence is enough when the alternating step is not too large. Note that in the encoder phase, the model can only adjust the RNN encoder-decoders to control the KL divergence, the generating parameters for latent variables are fixed.

\subsubsection{Scheduled Sampling Trick}
In the AE phase, we also find it useful to initially use the ground-truth encoding $\tilde{z}$ then gradually change to noisier CVAE output $z$. We apply the scheduled sampling strategy proposed in \cite{bengio2015scheduled}. Before decoding, a coin is flipped to decide whether to feed the real hidden vector $\tilde{z}$ or the noisy $z$. In the beginning, to make it easy, we mostly pick the real $\tilde{z}$. As the training proceeds, we gradually improve the difficulty by increasing the chance of selecting noisy $z$ until finally all inputs are replaced with the $z$. We decide the chance of selecting the real $\tilde{z}$ with a linear decay function as:
\begin{equation}
\label{eq: sampling}
p=\max (1-\frac{i}{k},0)
\end{equation}
$i$ is the step number and $k$ is a constant controlling the decaying speed. Other decaying functions are also applicable like exponential decay or inverse sigmoid decay.
\subsubsection{Training Process}
Our model contains a CVAE phase and an AE phase. These two phases are trained iteratively until an equilibrium is achieved. 

In the CVAE phase, A sample $\tilde{z}$ is obtained from the AE by transforming dialogue texts into a continuous embedding and is used as a target for the maximum likelihood training of the CVAE. We assume the generative model $p_\theta(z|\epsilon,c)=\mathcal{N}(\tilde{z},I)$, the loss function is:
\begin{align}
\begin{split}
\min_{\phi} \text{KL}(q_\phi (\epsilon|\tilde{z},c) \| p_\phi &(\epsilon|c))+\frac{1}{2}\mathbb{E}_{q_\phi (\epsilon|\tilde{z},c)}||g_{\phi}(\epsilon)-\tilde{z}||^2_2;\\
&\tilde{z}=f_\theta(x)
\end{split}
\end{align}
$f_\theta$ is the RNN encoder and is fixed as part of the AE module during training. 

In the AE phase, An observation x is sampled from the training data and fed into the transform function to get a continuous vector representation $\tilde{z}=f_\theta(x)$. The corresponding latent variable $z$ is sampled from the posterior distribution $q_\phi(z|\tilde{z},c)$ provided by the CVAE part. The sampled latent variable $z$, together with $x$, forms a target for training the AE. The objective function is:
\begin{align}
\label{eq: m1}
\begin{split}
\min_{\theta}\: &\max(\alpha, \text{KL}(q_\phi (\epsilon|\tilde{z},c) \| p_\phi (\epsilon|c)) )\\&-\mathbb{E}_{q_\phi (z|\tilde{z},c)}[log(p_\theta(x|z,c))];\\
&\tilde{z}=f_\theta(x),z=(1-p)g_\phi(\epsilon)+p\tilde{z}
\end{split}
\end{align}
The first item is used to control KL divergence in a reasonable range such that the transformed $z$ can be more easily modelled by the CVAE phase. $\alpha$ can be used to adjust the leverage between the reconstruction loss and KL divergence, where a lower $\alpha$ value will lead to a lower KL divergence in the end. $p$ is the keeping rate defined in Equation. \ref{eq: sampling}. The detailed architecture is depicted in Figure \ref{fig:model}. We refer to this framework as collaborative VED where the AE and CVAE phase collaborate with each other to achieve a better generating performance.

\subsubsection{Model Summary}In summary, we replace the GAN phase of AED with a CVAE alternative. The output of the CVAE part are latent variables, which can represent a much broader distribution family than mean-field Gaussian. As CVAE is in theory less accurate than GAN because it needs to approximate the real posterior, we leverage the more powerful RNN encoder-decoders. In the AE phase, they should autoencode utterenaces to make the real posterior easily representable by the CVAE part.
\section{Experiments}
We conduct our experiments on two dialogue datasets: Dailydialog~\cite{yanrandaily} and Switchboard~\cite{godfreyswitchboard}. Dailydialog contains 13118 daily conversations under ten different topics. This dataset is crawled from various websites for English learner to practice English in daily life. Statics show that the speaker turns are roughly 8, and the average tokens per utterance is about 15, which are appropriate for training dialog models. Switchboard has 2400 two-sided telephone conversations under 70 specified topics with manually transcribed speech and alignment. Compared with Dailydialog, the turn of every dialogue is much longer and the subject is more disperse. These two datasets are randomly separated into training/validation/test sets with the ratio of 10:1:1.

\subsection{Models and Training Procedures}
For comparison, we also implemented the hred model (seq2seq model with hierarchical RNN encoders), which is the basis of VHRED. Latent variable models are trained by standard KL-annealing with different weights~\cite{bowman2016generating,higgins2017beta}, with additional BOW loss~\cite{zhao2017learning,semeniuta2017hybrid}, word drop-out~\cite{bowman2016generating}, free bits~\cite{kingma2016improved} and our collaborative VED (CO) with the scheduled sampling trick (SS). For our framework, we use the encoder RNN as the transformation function $f_\theta(x)$. We tuned the parameters on the validation set and measure the performance on the test set. In all experiments, the letters are all transformed to the lower-case, the vocabulary size was set as 20,000 and all the OOV words were mapped to a special token $<$unk$>$. We set word embeddings to size of 300 and initialized them with Word2Vec embeddings trained on the Google News Corpus.  The first, second-layer encoder and decoder RNN in the following experiments are single-layer GRU with 512, 1024 and 512 hidden neurons. The dimension of latent variables is set to 512. The batch size is 128 and we fix the learning rate as 0.0002 for all models. Our framework is trained epochwise by alternatively training the CVAE and DAE part. The probability estimators for VAE are 2-layer feedforward neural networks. At test time, we output the most likely responses using beam search with beam size set to 5~\cite{graves2012sequence} and $<$unk$>$ tokens were prevented from being generated. We implemented all the models with the open-sourced Python library Tensorflow~\cite{abadi2016tensorflow} and optimized using the Adam optimizer~\cite{kingma2014adam}. Dialogs are cut into set of slices with each slice containing 80 words then fed into the GPU memory.
\subsection{Metric-based Evaluation}
We compare our model with the basic HRED and several current approaches including KL-annealing (KLA), word drop-out (DO), free-bits (FB) and bag-of-words loss (BOW). The details are summarized in Table \ref{tab: metric} and \ref{tab: embedding}. For KLA, we initialize the weight with 0 and gradually increase to 1 in the first 12000 or 25000 training steps for Dailydialog and Switchboard respectively. The word drop-out rate is fixed to 25\%. Words are dropped out only in the training step. We set the reserved space for every dimension as 0.01 in free bits (FB) and also try reserving 5 bits for the whole dimension space (FB-all). We use an $\alpha$ value 5 for our collaborative model (CO) and set the scheduled sampling (SS) weight $k=2500$ or 5000 for Dailydialog or Switchboard. We also experiment with jointly training the AE and CVAE part in our model and report the results.
\begin{table}[htbp!]
\centering
\caption{Metric Results, left: Dailydialog, right: switchboard}
\label{tab: metric}
\begin{tabular}{llll}
 \textbf{Model}&\textbf{PPL}  &\textbf{KL}  &\textbf{NLL}   \\
 \hline
 HRED&43.4$\vert$48.3  &0.00$\vert$0.00  &229.1$\vert$355.6    \\
 KLA&31.8$\vert$44.5  &4.90$\vert$4.36  &225.0$\vert$331.6  \\
 KLA+DO&29.8$\vert$40.1  &3.80$\vert$4.48  &223.9$\vert$317.0   \\
 KLA+BOW&26.8$\vert$30.9&12.8$\vert$8.92&247.3$\vert$321.1\\
 FB&41.7$\vert$32.1&\textbf{3.34}$\vert$\textbf{3.90}&239.0$\vert$322.7\\
 FB-all&29.4$\vert$21.7&5.01$\vert$4.97&226.1$\vert$308.2\\
 \hline
 CO&26.1$\vert$36.5&4.90$\vert$4.94&223.6$\vert$289.7\\
 CO+DO&25.1$\vert$34.4&5.01$\vert$4.93&218.7$\vert$273.4\\
 \textbf{CO+SS}&\textbf{23.8}$\vert$\textbf{31.8}&4.92$\vert$4.93&\textbf{213.2}$\vert$\textbf{273.4}\\
 CO+SS(joint)&28.5$\vert$39.6&5.16$\vert$5.02&224.3$\vert$301.3\\
\end{tabular}
\end{table}
Table \ref{tab: metric} measured the perplexity (PPL), KL divergence (KL) and negative log-likelihood (NLL). NLL is averaged over all the 80-word slices within every batch. For latent variable models, NLL is computed as the ELBO, which is the lower bound of the real NLL.

As can be seen, our model CO+SS achieves the lowest NLL over both datasets. The Schedule Sampling (SS) strategy significantly helps brings down the NLL. Word drop-out (DO), though weakening the RNN decoder, improved the performance when combined with both KLA and CO, which verified the assumption that DO can function as a smoothing technique in neural network language models~\cite{xie2017data}. KLA itself needs early stop, otherwise the KL divergence will vanish once the weight increases to 1. BOW avoids the KL-vanishing problem, but the overall performance will significantly decrease because adding an additional loss in theory leads to a biased result for latent variables. BOW information is encoded into the latent variable, but it prevents the decoder from stably learning the word order pattern in the training step thus sacrifices the NLL performance. FB-all performs much better than FB, which suggests most important information is concentrated on a few dimensions. Equally reserving space for every dimension is not suitable. Finally, we also testified the necessity of iteratively training our model. Jointly training the model brings recession on both the perplexity and KL divergence on the two datasets.

\begin{figure}[!ht]
\centering
\includegraphics[width=8cm,height=4cm]{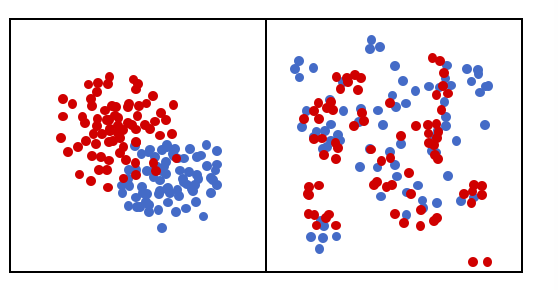}
\caption{T-SNE visualization of sampled latent variables. left: VHRED, right: CO+SS. Red dots correspond to samples from prior distribution, while the blue dots correspond to samples from posterior distribution}
\label{fig:scatter}
\end{figure}

Figure. \ref{fig:scatter} visualizes the latent variables drawn from VHRED and our framework. We randomly pick a dialogue context ``I'd like to invite you to dinner tonight , do you have time ?" and apply the information retrieval based method to gather 10 responses with similar context from the corpus. All the 10 responses are verified by humans as appropriate ones, which span over different possibilities like ``Thank you for your invitation. ", ``Don't be silly . Let's go Dutch ." and ``Are you asking me for a date ? ". For each response, 10 samples are drawn from the posterior latent variable distribution, which forms 100 posterior latent variable samples (blue dots) in total. Likewise, 100 samples are drawn from the prior latent variable distribution (dots) given only the dialogue context. The visualization clearly indicates the superiority of our framework in modelling more flexible prior and posterior latent variable distributions. In the VHRED model, both the prior and posterior distributions are limited uni-modal Gaussians with only a little overlap. In our framework, the distributions are more diverse and samples from the prior and posterior distribution share more overlap with each other.

\begin{table}[htbp!]
\centering
\caption{Embedding Results, left: dailydialog, right: switchboard}
\label{tab: embedding}
\begin{tabular}{llll}
 \textbf{Model}&\textbf{Average}  &\textbf{Greedy}  &\textbf{Extrema}   \\
 \hline
 HRED&0.463$\vert$0.334  &0.445$\vert$\textbf{0.399}  &0.356$\vert$0.280    \\
 KLA&0.442$\vert$0.317  &0.436$\vert$0.327  &0.327$\vert$0.267   \\
 KLA+DO&0.458$\vert$0.325  &0.461$\vert$0.341  &0.378$\vert$0.283  \\
 KLA+BOW&0.475$\vert$0.340&0.459$\vert$0.352&0.386$\vert$0.302\\
 FB&0.423$\vert$0.336&0.414$\vert$0.348&0.349$\vert$0.318\\
 FB-all&0.429$\vert$0.341&0.439$\vert$0.352&0.357$\vert$0.325\\
 \hline
 CO&0.465$\vert$0.377&0.465$\vert$0.381&0.394$\vert$0.331\\
 CO+DO&0.489$\vert$0.385&0.471$\vert$0.379&0.397$\vert$0.337\\
 \textbf{CO+SS}&\textbf{0.539}$\vert$\textbf{0.392}&\textbf{0.477}$\vert$0.394&\textbf{0.443}$\vert$\textbf{0.340}\\
  CO+SS(joint)&0.420$\vert$0.347&0.452$\vert$0.360&0.351$\vert$0.308\\
\end{tabular}
\end{table}

Table \ref{tab: embedding} reports the results of the embedding-based topic similarity: Embedding Average (Average), Embedding Extrema (Extrema) and Embedding Greedy
(Greedy)~\cite{liu2016not}. Unlike the NLL, who measures the token-level match, these embedding-based metrics map responses to a vector space and compute the cosine similarly with golden answers, which can to a large extent measure the sentence-level semantic similarity.

We can see our model still achieved the highest topic similarity according to all the three metrics. This suggests our model can bring improvement for both token-level coherence and sentence-level topic match. BOW, though not good at the NLL metric, performed remarkably well on this metric, which implies BOW is beneficial for the decoder to generate the correct high-level meaning but fails to transform the meaning to a fluent sentence. In contrast, FB has a relative lower on-topic similarity score compared with its performance on the token-level likelihood.  
\subsection{Human Evaluation}
\begin{table*}
\caption{{\label{tab: examples}}Examples of context-response pairs for the neural network models. \_\_eou\_\_ denotes end-of-utterance and indicates the start of a new turn.}
\centering
\begin{tabular}{p{60mm}|p{110mm}}
  \textbf{Context} & \textbf{Response}\\
  \hline
  The taxi drivers are on strike again . \_\_eou\_\_ What for ? \_\_eou\_\_ They want the government to reduce the price of the gasoline . \_\_eou\_\_ &  \textbf{KLA:} \ all right . i'll take it . \newline \textbf{KLA+BOW:} \ well , how much is the price of the price price ? \newline \textbf{CO-SS:} \ that's good . i want to buy a new car .  \newline \textbf{Gold:} \ It is really a hot potato .  \\ \hline
   I'd like you to do me a favor . \_\_eou\_\_ What is it ? \_\_eou\_\_ Can you mail this at the post office for me ? I have to go to the doctor for an appointment and I don't have enough time to go to the post office . \_\_eou\_\_ &  \textbf{KLA:} \ i'm sorry , but i'm not sure . i don't want to have a look at the moment . \newline \textbf{KLA+BOW:} \ oh , i'm sorry , but it's a pleasure to be in my room and again-b .  \newline \textbf{CO+SS:} \ ok , i will . do you have any other questions about the company ?   \newline \textbf{Gold:} \ Sure . And I'd like you to get some cold medicine for me when you go to the doctor's .  \\ \hline
   I heard you've found a new job . \_\_eou\_\_ Yes , I'm now working in a company . \_\_eou\_\_ So you're quite happy with your new job ? \_\_eou\_\_ &  \textbf{KLA:} \ to be honest , but i'm not familiar with my friends .  \newline \textbf{KLA+BOW:} \ i'd like to . but i would like my mind .  \newline \textbf{CO+SS:} \ not really , but how about you , sue ?   \newline \textbf{Gold:} \ Right . I enjoy what I'm doing .  \\ \hline
\end{tabular}
\end{table*}

The accurate evaluation of dialogue systems is an open problem. To validate the previous metric-based results, we further conduct a human evaluation on several models. We randomly sampled 100 context from the test corpus and apply 6 different models to generate the best response with beam search. The evaluation is conducted only on the Dailydialog corpus since it is closer to our daily conversation and easier for humans to make the judgement. All the generated responses, together with the dialogue context, are then randomly shuffled and judged on the crowdsourcing website CrowdFlower. People are asked to judge the plausibility of the generated response by giving a binary score in three aspects: grammaticality, coherence with the dialogue context and diversity (ensure the response is not a dull sentence). 54 people are finally involved in evaluating the total 600 responses, each is judged by 3 different people and the score agreed by most people is adopted. We set each person can judge at most 50 responses and filter by manually-set test questions.

The results shows that our model generates highly fluent sentences compared to other approaches. KLA+BOW, as expected, receives the lowest score on fluency. Our model also achieves relative good scores on coherence and diversity, implying novel responses related to the conversation topic can be generated by our model. However, we notice the human evaluation is rather subjective and not reliable enough. If a sentence is influent, humans tend to reject it though the topic might be coherent and the content might be diverse. It is difficult to give an objective score separately for all the three aspects. We can see models with lower scores on fluency normally also receive lower scores on the other two fields like KLA+BOW and FB-all. Therefore, we consider this evaluation only as a complement to the metric-based results, indicating that humans agree with the generations of our models more than with the others.
\begin{table}[htbp!]
\centering
\caption{\label{tab:human scores}Human Judgements for models trained on Dailydialog corpus, F refers to fluent, C refers to coherence and  D refers to diversity.}
\label{tab: human}
\begin{tabular}{llll}
 \textbf{Model}&\textbf{F(\%)}  &\textbf{C(\%)}  &\textbf{D(\%)}   \\
 \hline
 KLA&76  &35  &50   \\
 KLA+DO&80  &41  &\textbf{57}   \\
 KLA+BOW&70&36&48\\
 FB-all&74&29&34\\
 \hline
 CO+DO&82&\textbf{49}&54\\
 \textbf{CO+SS}&\textbf{89}&44&51\\
\end{tabular}

\end{table}

Table \ref{tab: examples} shows exampled generated responses. We can see the our improved collaborative VED model with scheduled sampling can more accurately identify the topic and generate more coherent responses. Standard KL-annealing tends to generate smooth sentences but irrelevant to the context. Imposing an additional BOW loss can increase the probability of correctly capturing the main topic, but the generated responses are sometimes grammatically wrong, as also has been shown from the metric-based results. In the first example, the context is about taxi drivers' request for reducing gasoline price, the response from KLA is a fluent natural sentence but not closely related to the context. Model KLA+BOW starts with a reasonable beginning but ends up with influent continuations. Though influent, KLA+BOW model does capture the main topic about price, indicating it can successfully predict the order-insensitive bag of words but fail to establish a natural sentence. In contrast, our model is not only a fluent sentence, but also close to the topic. More importantly, it brings some new information ``I want to buy a new car" and is helpful to an interactive conversation. Similar conditions can be seen in the other two examples.
\section{Conclusion}

Variational encoder-decoders and recurrent neural networks are powerful in representation learning and natural language processing respectively. Though recently quite a few work has started to apply them on dialogue generation, the training process is still unstable and the performance is hard to be guaranteed. In this work, we thoroughly analyze the reason of the training difficulty and compare different current approaches, then propose a new framework that allows effectively combining these two structures in dialogue generation. We split the whole structure into two parts for more flexible prior and posterior latent variable distributions. The training process is simple, efficient and scales well to large datasets.

We demonstrate the superiority of our model over other popular methods on two dialogue corpus. Experiments show that our model samples latent variables with more flexible distributions without sacrificing recurrent neural network's capability of synthesizing coherent sentences. Without losing generality, our model should be able to apply on any se2seq tasks, which we leave for future work.
\subsubsection{Acknowledgements} Xiaoyu Shen is supported by IMPRS-CS fellowship. The work is partially funded by DFG collaborative research center SFB 1102 and the National Natural Science of China under Grant No. 61602451.
\bibliography{aaai}
\bibliographystyle{aaai}
\end{document}